# Reinforcement Learning in Education: A Multi-Armed Bandit Approach

Herkulaas MvE Combrink [1] [0000-0001-7741-3418], Vukosi Marivate [1] [0000-0002-6731-6267] and Benjamin Rosman [2] [0000-0002-0284-4114]

[1] Department of Computer Science, University of Pretoria, South Africa
[2] School of Computer Science and Applied Mathematics, University of the Witwatersrand, Johannesburg, South Africa
u29191051@tuks.co.za

**Abstract.** Advances in reinforcement learning research have demonstrated the ways in which different agent-based models can learn how to optimally perform a task within a given environment. Reinforcement leaning solves unsupervised problems where agents move through a state-action-reward loop to maximize the overall reward for the agent, which in turn optimizes the solving of a specific problem in a given environment. However, these algorithms are designed based on our understanding of actions that should be taken in a real-world environment to solve a specific problem. One such problem is the ability to identify, recommend and execute an action within a system where the users are the subject, such as in education. In recent years, the use of blended learning approaches integrating face-to-face learning with online learning in the education context, has increased. Additionally, online platforms used for education require the automation of certain functions such as the identification, recommendation or execution of actions that can benefit the user, in this sense, the student or learner. As promising as these scientific advances are, there is still a need to conduct research in a variety of different areas to ensure the successful deployment of these agents within education systems. Therefore, the aim of this study was to contextualise and simulate the cumulative reward within an environment for an intervention recommendation problem in the education context.

## 1 Introduction

### 1.1 Recommender systems

The fourth industrial revolution (4IR) is a disruption and augmentation to real-time processes interwoven with the digital domain [1]. In the context of South Africa, innovation in 4IR technologies is still in its inception as compared to first world countries [2]. Furthermore, 4IR disruptions are mostly observed within industrial mechanical processes, but there are certain domains that might not have benefitted from the full adoption of these disruptive 4IR technologies, such as in first world education [3]. In the context of education, great strides have been made in terms of developing information and communication technologies (ICTs) that can enable the storage and retrieval of learning material such as learning management systems, through to ICTs that can connect people on digital platforms from anywhere in the world [4]. As the software and autonomous learning components within systems increase, there is a growing need to understand the adaptive automation elements within such a system [5]. Adaptive automation as a concept refers to the locus of control between humans and machines, and where these boundaries reside. An example of an adaptive automation theory used was developed by D'Addona *et al.* (2018). In this framework, where automation is needed, and human decision making is outlined is illustrated using an industrial process (Fig. 1) [6]. Most adaptive automation frameworks are used in industrial processes, but these frameworks can also be applied to other contexts, such as education [3].

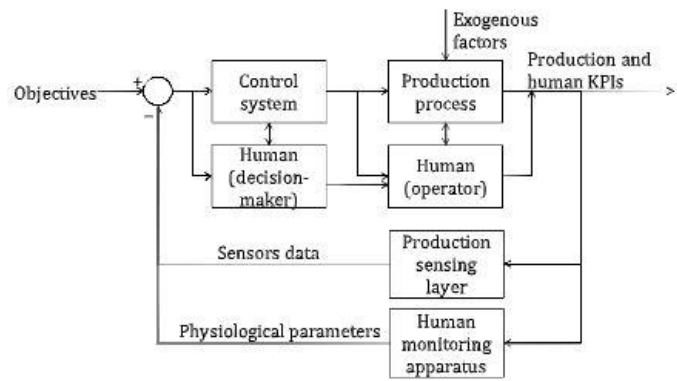

Fig. 1: Framework for human-in-the-loop factory adaptive automation[1]

However, there is still unexplored potential regarding the integration of autonomous learning systems in all spheres of the education domain [7]. One such example is recommender systems [8]. A recommender system is a dynamic machine learning filtering algorithm that uses data and a context to recommend an outcome to a user within a specific system [9, 10]. For example, some of these algorithms include collaborative filtering (context about the user to make the recommendation) to content-based filtering (context about the item that is being recommended) [11, 12]. Companies such as Google and Netflix have made use of these kinds of algorithms to recommend content to users browsing for content on these platforms [13, 14]. As promising as these algorithms are, they also have pitfalls in their implementation.

---

[1] source: https://www.sciencedirect.com/science/article/pii/S00078 specifix50618301471

In the education context, collaborative filtering relies on the average student's (user) input to recommend an intervention (the item). This approach runs the risk of applying generic and generalisable interventions to students that require nuanced approaches [15]. On the other hand, content-based filtering places an emphasis on the specific student's (user) own experience of the intervention (item) – but unfortunately this does not scale well in the South African context as students, their cultures and contexts differ [16]. In recent years, a hybrid approach has been adopted, using a combination of elements from both of these to make a sufficient recommendation [17].

As great as these systems are, they too fall short with reference to a few fundamental problems. Firstly, recommender systems suffer from the cold start problem [18]. The cold start problem refers to a situation that occurs when no information is available about a user or the item in the system (student or the required intervention). There are three types of cold start problems that can be classified based on where in the system the data are missing [19]. Type one occurs when a new student (user) enters the system without any prior knowledge on the user, type two when a new intervention (item) enters the system, and type three when the system starts for the first time. This means that recommender systems that are implemented suffer from this problem when the system starts for the first time. Secondly, recommender systems suffer from issues related to data sparsity. This means that not all the relevant information is collected about all the users within a given system. As a result, mislabelling of the interventions (items) is possible in the education context, given that sparsity challenges such as missing data about a specific item or set of users might occur. Thirdly, scalability is a major concern with these systems. The issue of scalability arises on the premise of whether or not the system will be able to cope as the number of students and interventions grows [20]. In studies about recommender systems, it was found that the moment a certain threshold or sample size is reached (depending on the algorithm and system) the results might not be desirable [21]. The fourth major problem relates to the lack of data within a given system [22]. If the system does not have at least enough data from the latent variables used associated with the different users, then the system cannot effectively make any recommendations. The final set of problems relate to a change in data and user preferences. In the education context, students (users) have different needs, and therefore require different interventions (items) to be recommended to them. As time continues, the needs of the same student might change, and as a result, the recommendation should subsequently support this change as well (Fig. 2).

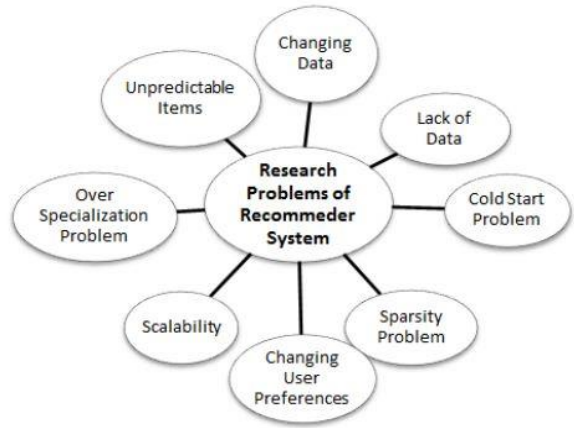

Fig. 2: Research problems associated with recommender systems[2]

In addition to this, learning design to recommend a specific intervention for a specific student based on their specific needs is a common problem with recommender systems. It is a common problem because the challenge lies in when to recommend what intervention would fit the needs of the student. Not all students would require a recommendation at the same time, and the same recommendation would not have the same impact for all students. Although recommender systems show a lot of promise to provide a solution in a variety of contexts in higher education, unfortunately, there are several technical, philosophical, and pedagogical hurdles that need to be overcome before a scalable solution to recommending interventions to students can be identified. Finding new information about students in real-time might not be a viable solution because these systems might only learn what was valuable to implement retrospectively [23]. To learn in real-time and work with data that is novel and not collected prior, unsupervised learning might be a possible solution. In machine learning, the algorithms' function using context learned from structured data [24]. In unsupervised learning, the approach fundamentally differs from supervised learning as an agent, which is an autonomous decision maker in the form of an algorithm, needs to make choices based on actions and be rewarded in relation to the choices it made within a specific environment [25]. This type of approach is known as reinforcement learning, which is a computer simulation based on principles similar to those found in classical conditioning but applied in the context of a decision-making framework [26].

---

[2] source: https://iopscience.iop.org/article/10.1088/1742-6596/1717/1/012002

## 1.2 Multi-Armed Bandits and Markov Decision Processes

Let us propose that an agent is an autonomous decision maker within a specific context. The multi-armed bandit (MAB) problem is a type of reinforcement learning algorithm where the agent needs to make one of two choices per 'arm' in the system [27]. The purpose of the MAB is to solve the problem of choosing the arm within a given context that gives the overall highest rewards [28]. The challenge with traditional MAB problems was that it was applied to very specific situations where the fundamental scenario remained the same, leaving no room for a situation where the actions taken have an influence on future actions [29]. As a result of these shortcomings, Markov Decision Processes (MDP) were introduced to MAB to simulate more complex situations where the environment takes into account the influence current actions have on determining the best future action [30].

This type of autonomous decision making can be represented as a tuple, whereby the MDP state-action-reward loop can be denoted by $\mu = (S, A, T, R, \gamma)$ [31]. In this tuple, *S* represents conventional states, *A* signifies the finite actions to be taken by the agent, T denotes the transition from one state to another, *R* being the reward function, and $\gamma$ the specific discount factor within the MDP. Within this tuple, the *T*, *R*, and $\gamma$ are interdependent on conditions for them to function within the MDP. The T element within the tuple can be represented as the placement from $S_t$ to $S_{t+1}$ given a certain *A* so that $T: S_t$ x *A* x $S_{t+1} \rightarrow [0, 1]$ from state 0 to 1 can be denoted as a function of *T (s, a, s')* representing the probability of transitioning from state *s* to state *s'*, given a specific *A*. The *R* element within the tuple can be denoted as *R (s, a, s')* for the same state action pairs. The context of the MDP is within a specific set of task instances, which can be denoted as *X*. This representation allows for the $\gamma$ to be denoted as $\gamma \in [0, 1]$ given a set of task instances within the MDP space *(M)*. Within the MDP, all state action pairs, including the *T (s, a, s')* and *R (s, a, s')* will be referred to as episodes, bound by a specific time context. The probability distribution of taking a specific action based on certain states is called a policy, and is denoted by $\pi$. When a policy is tracked over several episodes, the cumulative discounted reward can be calculated as a function of the *R* at a specific *T* for each state-action-reward loop. Given these annotations, the problem that an agent is trying to solve is to choose an *A*, between different *S*, within several instances of *T*, to maximize the cumulative reward based on the highest *R* for each episode [31].

Reinforcement learning does not suffer from the cold start problem and depending on how the reinforcement learning environment is designed, the context from which the agent takes an action and the types of interventions such a system could recommend, are fully customizable and do not suffer from the change in data problem [32, 33]. Lastly, reinforcement learning can transfer knowledge and context learned based on the specific problems that were solved [29 – 33]. Given this, reinforcement learning has great potential as an alternative for student intervention recommendations [34]. Therefore, the aim of this study was to contextualise and simulate the cumulative reward within an environment for an intervention recommendation problem in the education context.

## 2 Methods

### 2.1 MAB Student Intervention Recommendation Framework

To build a system for student intervention recommendations, the environment must be custom made to simulate specific actions taken by the agent and assess how the overall cumulative reward will be impacted by decision making and context in the education simulation. What we should not do within this environment is decide on the specific interventions as this is reserved for domain specific experts. In addition to this, the autonomy an agent has over deciding what type of intervention is best suited for a specific student will be stratified based on hypothetical interventions that are recommended by the agent. In this simulated environment, there should be broad enough categorization of the outcome of a specific intervention, assuming that the emphasis is not on the details of the specific intervention, but rather what constitutes success in implementing such an intervention. For example, if intervention X is recommended to the student and the student follows through on this, then the student has a certain percentage of passing based on the intervention proposed. In other words, the agent will make a choice based on the impact the recommendation will have. Given that the agent can make mistakes (choosing a hypothetical intervention that will not assist the student), the experiments need to include scenarios that illustrate the impact of making mistakes within a given context as well. Furthermore, it is pivotal that each specific student is contextualized in the framework of this problem. To achieve this, an assumption is made that there is a process prior to the MAB problem which correctly categorizes a student based on the problems the students face. The emphasis of this work is on the intervention recommendation component of such a complex system, and not the identification of the plethora of challenges that may exist within this context or how to correctly classify a student based on the interventions they need. Lastly, we assume that each student is a state, the action is the recommendation (or how effective the recommendation is in the context of this simulation), and the reward is assumed (although in a real-world system, this will be learned by the agent). The assumption of the reward is primarily for the purpose of evaluation so that the decision the agent made can be tested against the reward functions for those specific situations (Fig. 3).

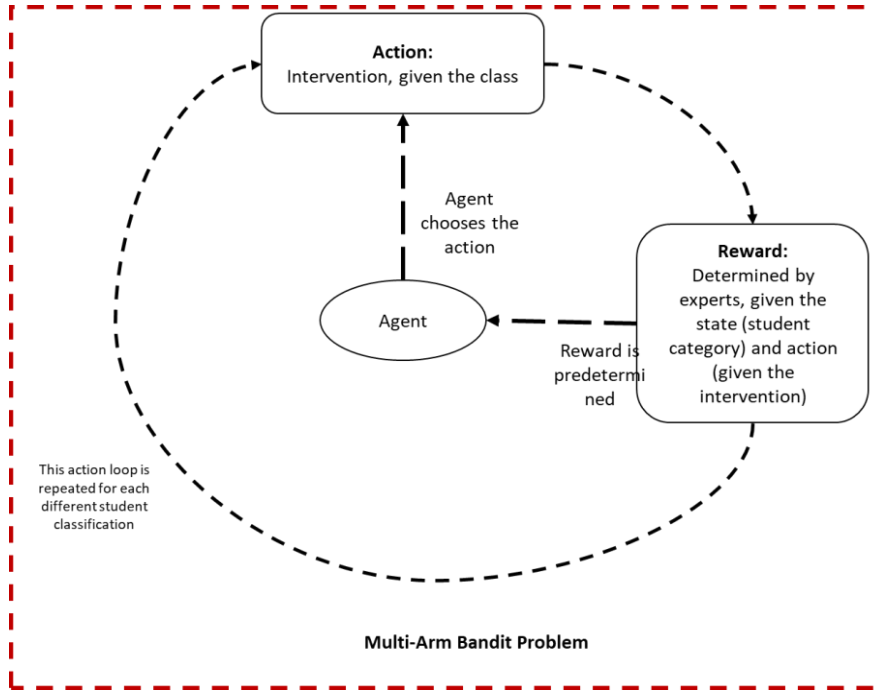

Fig. 3: Schematic representation of the student intervention MAB environment

### 2.2 Simulation Parameters

We are basing our assumptions on four different reward probabilities given to the environments, based on the type of student classification made. There are hundreds of different student classifications that can be given, each with their own unique interventions. We are only concerned with whether the correct recommendation is made, and not what the recommendation is. We assume that the students were correctly classified. In other words, we are assuming that the classification was perfect for the given student. The agent can also take one of four actions. The first action signifies an incorrect intervention; action two is an intervention with an unknown consequence; action three represents the ideal intervention; and action four represents an intervention that may work, but not for all instances.

We are assuming that all the intervention distributions per class are labelled by experts, and that the interventions are perfect. For our experiments, four different categories of student were used. These four groups or classes of student represent conditions for interventions that are given within education, dependent on the student needs and the problems they face [35 – 37]. The first class is a student that will pass, no matter the intervention. The second category is a student that will fail but will only pass if the correct intervention is given and stand a 70% chance of passing if the intervention that may work is recommended, and 50% if an incorrect intervention or an intervention with an unknown consequence is given. The third student will fail but will only pass if the correct intervention is given and stand a 50% chance of passing if the intervention that may work is recommended, and 25% chance that an unknown intervention is recommended. The fourth and last student is a student that will fail and will only pass 50% of the time if the correct intervention is given.

On average, the majority of the students who met the entry requirements for a specific degree need minimal to no input to pass their qualification. The exact number has not been agreed upon yet as a domain standard, but the majority of students that will pass, without intervention – roughly 55%. We will classify these as category 1 students [38]. In our example, approximately 20% of students are likely to pass if a specific or specialized intervention is given, and without such an intervention, they are 50% likely to fail, we will constitute this as category 2 [35 – 39]. Furthermore, 10 – 20% of students will fail within a given system if the right intervention is not given, and if the right intervention is given, are still likely to fail. We will group this as category 3. Additionally, 5 – 10% of students will fail within a system due to reasons outside the control of the system. If an intervention is given to these students, they are still likely to fail with little to no chance of success. We will group this as category 4 [35 – 42]. All the action pairs and their associations with the different categories are summarized in Table 1 below.

Table 1: Summary of simulation parameters in the experiment

| Name of Category | The % distribution within cohort | Likelihood to pass for recommendation 1 | Likelihood to pass for recommendation 2 | Likelihood to pass for recommendation 3 | Likelihood to pass for recommendation 4 |
|---|---|---|---|---|---|
| Category 1 | ~55% of cohort | 100% | 100% | 100% | 100% |
| Category 2 | ~20% of cohort | 50% | 50% | 100% | 70% |
| Category 3 | ~10% of cohort | 0% | 0% | 100% | 25% |
| Category 4 | ~5% of cohort | 0% | 0% | 50% | 0% |

Within the confines of the experiment, each category of student will represent its own environment, that is, a simulation will be performed for each environment as outlined in Fig. 3.

## 2.3 MAB Algorithms

We used three different algorithms to simulate their impact on the proposed education environment. The first algorithm is a random agent. That is, an agent that takes a random $A$ based on the available actions. The purpose of including this was to show the impact of taking random non-informed actions in the form of recommending random interventions within this environment. The second Algorithm was an epsilon greedy algorithm. The epsilon greedy algorithm makes decisions based on the trade-off between exploration and exploitation. Exploration in this instance can be defined as taking an action with an unknown return, and exploitation can be defined as making a decision based on known circumstances and known outcomes. Epsilon can be characterized as the percentage of times decisions are made from an exploration perspective. In the context of this study, epsilon greedy was explored because there are risks associated with exploration when dealing with human subjects. Within the design of the algorithm, epsilon was set to 1%, that is, to explore 1% of the decision made.

The third and final algorithm used in the experimentation was upper confidence bound (UCB). This algorithm works by implementing an exploration and exploitation approach, but unlike epsilon greedy, the exploration-exploitation trade-off is updated as the agent learns more from its environment. This algorithm was included as UCB is an illustration of an agent that has been used in other autonomous learning systems. There are several algorithms that have been used to solve a MAB problem such as temporal difference learning and Bayesian Policy reuse, but for the purpose of this paper, the emphasis is on the evaluation of the cumulative reward within an environment simulating an education student intervention recommendation problem, and not which specific algorithm works best for this environment. To visualise the data from the experiments, the distribution of the recommended interventions as well as the cumulated rewards over the number of episodes were shown. As mentioned prior, each episode is representative of a specific class of student that requires a specific intervention. Within each experiment, each of the four types of categories of student were represented within their own environment, that is, for each of the four categories of student, a different environment was used. Each simulation was repeated 10,000 times, and a confidence interval of 95% was adjusted around the results. Lastly, the distribution of the choices made per environment were illustrated to show how the various agents made decisions across the distributions of the types of interventions that were recommended (Table 1).

## 3    Results

The first experiment outlines what would happen if a non-smart agent is used to randomly make decisions (Figure 4). As expected, students who will pass no matter the intervention (Environment 1) reached the maximum rewards after 16 episodes (students). The reward function for students who would only pass 50% of the time if the correct intervention is given (Environment 4) had an average reward of ~10% (CI 8 – 12%). This illustration outlines that no matter the recommended intervention, students who will pass will not suffer from random interventions recommended to them, whereas the consequences of not recommending the right intervention to students at risk of failing is quite high.

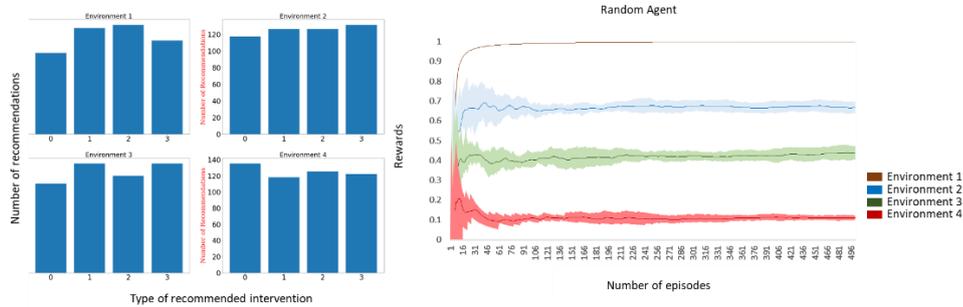

Fig. 4: Random agent recommending intervention simulation

The second experiment outlined what would happen if an agent had to implement decisions, given a certain level of exploration. In this experiment, the most important difference in Environment 4 was the average cumulative reward of ~20% (CI 5- 35%). In this simulation, students who require a specific intervention to pass (Environment 2 & Environment 3) overall had an ~80% (CI 75 – 90%) reward, indicating that a system implementing MAB will be able to provide a high level of accuracy at learning which intervention to implement for a specific class of student if the impact of the interventions recommended is known (Figure 5). What this approach also illustrated is that there are more risks associated with implementing an autonomous learning system that continuously explores as compared to an agent that randomly makes recommendations as seen with the lower limits of Environment 4 being 0% in several instances.

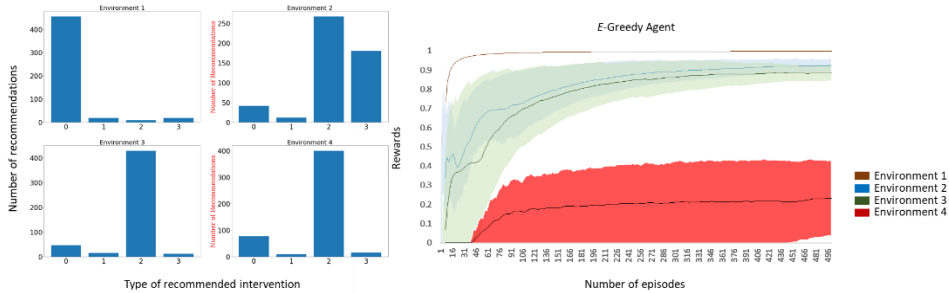

Fig. 5: Epsilon greedy agent recommending intervention simulation

In the last experiment, UCB was used as an illustration of how a better kind of autonomous algorithm can be used to make recommendations within an autonomous learning system. In this simulation, all the overall average rewards were higher than the previous experiments, with lower confidence intervals for each of the cumulative rewards in each environment (Figure 6).

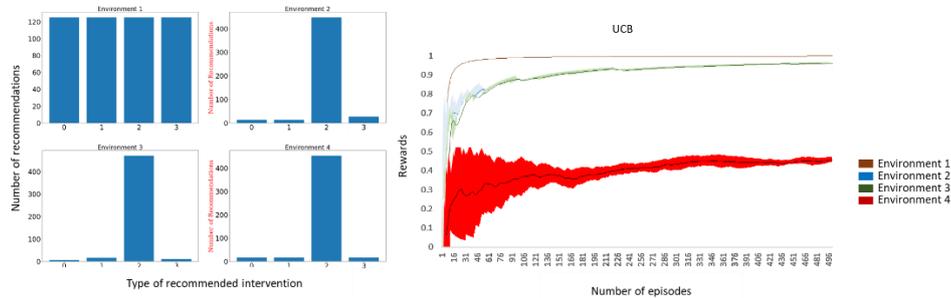

Fig. 6: UCB agent recommending intervention simulation

## 4   Discussion

Within institutions of higher learning, student retention and student dropout remain a problem [35]. Student retention refers to the students remaining within the institution despite their academic outcome the previous year, whereas student throughput refers to whether or not a student progresses through the system to make it to graduation. Several student support strategies and mechanisms have been explored to identify if they contribute toward the improvement of both student retention and throughput [36]. Some of these student support strategies include social support for students provided by the institution, academic support strategies facilitated by the institution, and high impact practices fostering a sense of learning in the students implemented by the institution [36, 38, 39, 41]. The types of interventions that best fit the challenge a student faces is dependent on a variety of different factors, such as the needs of the student, which interventions are available by the institution, and which intervention is the best fit to aid in solving the problem [41, 42].

In this study, several abstract examples were given to simulate how different agents within a MAB problem will accumulate rewards over time. If the student will pass, no matter the outcome, then any system can implement an intervention. On the other hand, if an intervention only has a 50% success rate, if and only if the correct intervention was implemented to the correct student, then the algorithm will not learn beyond the current human capacity to solve the problem. What was also observed was that in certain instances, if the outcome of an intervention is not known, then there might be more risks associated with implementing a system that explores interventions as opposed to a system randomly recommending interventions. What these systems do really well, as illustrated in the simulations (Figures 3 – 6) is learn to what extent we can build a system, based on prior knowledge, and how well these systems can potentially recommend an intervention to a student if we have that knowledge and understanding. What these simulations also illustrated are the consequences of not knowing the extent to which interventions can benefit the students. The ethical dilemma is that if a system implements an intervention that could potentially cause harm to a student, is it the fault of the engineers implementing the systems, the policy makers approving the systems, or the lack of understanding experts in the education domain have regarding which interventions work to improve education outcomes? How will a combination of these issues be dealt with, and what fundamental policies need to be in place to assist decision making in this domain? That is why adaptive automation is still important because, at present, the knowledge gap is too wide to implement these systems autonomously, as the study illustrated. The purpose of systems autonomously implementing interventions based on the needs of students is vital but depends on our fundamental understanding of what interventions work for which students, and not how well the algorithms work at solving fundamentally unsolved human problems.

Although there are universal overlaps in schools of thought about the fundamental needs of a student, there are nuanced differences depending on the context from which the student functions [41]. These differences are affected by a variety of factors including socio-economic status, financial situation, home language, the institution's language of instruction, general support structures available to the student, as well as the general drive, persistence, and resilience of the student [43, 45]. Despite these complexities, there is a shift toward the use of systems, such as recommender systems for the higher education context. In the context of higher education recommendation, specifically with reference to student interventions, the purpose of the system is to identify the correct need of the student and recommend a solution to the student that could potentially motivate the student to pass if they are at risk of failing because the correct intervention may be recommended. However, as outlined prior, how well a system recommends a solution is still dependent on our understanding of the problem, and at present, the science is not of such a nature that the system can suggest a recommendation outside the scope of the designed environment and outside the scope of our current understanding of the problem.

What is further complicated about these types of systems is that they should comprise a series of different algorithms so that they can recommend the correct intervention to the correct problem. This makes student intervention systems difficult to implement because students are not homogeneous and the student population, cohort, and society as a whole change over time. The interventions available by the institution and the type of support strategies that will best fit the problem are dependent on the academics, support staff, and student support framework policies of the institution. Complications arise when scaling student support strategies because student needs are vast and complex. In addition to this, the cost and time required to drive real-time analytics to identify, recommend and evaluate the effectiveness of these interventions to students is high if these processes are driven by staff, even if there is a dedicated staff compliment fulfilling this function within a higher education institution [44, 45]. Therefore, despite the potential unintended consequences of these systems, they are significant and potentially instrumental in recommending interventions to students within education. Lastly, the simulations within the experiments were intended to show empirically the consequences if the outcomes of interventions are not known, as to illustrate a need for further research from experts who are working in the education domain.

## 5   Conclusion

It is essential to design smart systems within higher education institutions that are generalizable and can adapt to both similarities and differences between students. To create such a system, a series of analytical and computational models need to be evaluated as the fundamental process of establishing such a system should rely on a combination of education theory, human intervention, adaptability, and human-machine interactions embedded within a system. Such a system will require a series of analytics, machine learning, and potentially reinforcement learning algorithms, all applied within a student support framework and working in conjunction with people within the system to serve the students and truly assist them. Understanding how to implement interventions within complex systems is fundamentally important as technology alone cannot define the boundaries and scope of what interventions and what processes will be used and will not be used within the education context. The science of systems learning is expanding at a rapid rate, but this type of development will be truly beneficial if coupled with systems-based domain specific knowledge for it to be successful. As disruptive as 4IR might be, our fundamental understanding of concepts remains a top priority if we want autonomous learning to be implemented responsibly to benefit the education domain.

## 6   References

1. Coetzee, J., Neneh, B., Stemmet, K., Lamprecht, J., Motsitsi, C. and Sereeco, W., 2021. South African universities in a time of increasing disruption. *South African Journal of Economic and Management Sciences*, *24*(1), pp.1-12.
2. Rashied, N. and Bhamjee, M., 2020. Does the Global South Need to Decolonise the Fourth Industrial Revolution?. In *The Disruptive Fourth Industrial Revolution* (pp. 95-110). Springer, Cham.
3. Oke, A. and Fernandes, F.A.P., 2020. Innovations in teaching and learning: Exploring the perceptions of the education sector on the 4th industrial revolution (4IR). *Journal of Open Innovation: Technology, Market, and Complexity*, *6*(2), p.31.
4. Gamede, B.T., Ajani, O.A. and Afolabi, O.S., 2022. Exploring the Adoption and Usage of Learning Management System as Alternative for Curriculum

Delivery in South African Higher Education Institutions during COVID-19 Lockdown. *International Journal of Higher Education*, *11*(1), pp.71-84.

5. Bortolini, M., Faccio, M., Galizia, F.G., Gamberi, M. and Pilati, F., 2020. Design, engineering and testing of an innovative adaptive automation assembly system. *Assembly Automation*.
6. D'Addona, D.M., Bracco, F., Bettoni, A., Nishino, N., Carpanzano, E. and Bruzzone, A.A., 2018. Adaptive automation and human factors in manufacturing: An experimental assessment for a cognitive approach. *CIRP Annals*, *67*(1), pp.455-458.
7. Dwivedi, S. and Roshni, V.K., 2017, August. Recommender system for big data in education. In *2017 5th National Conference on E-Learning & E-Learning Technologies (ELELTECH)* (pp. 1-4). IEEE.
8. Obeid, C., Lahoud, I., El Khoury, H. and Champin, P.A., 2018, April. Ontology-based recommender system in higher education. In *Companion Proceedings of the The Web Conference 2018* (pp. 1031-1034).
9. Li, Q. and Kim, J., 2021. A deep learning-based course recommender system for sustainable development in education. *Applied Sciences*, *11*(19), p.8993.
10. Nouh, R.M., Lee, H.H., Lee, W.J. and Lee, J.D., 2019. A smart recommender based on hybrid learning methods for personal well-being services. *Sensors*, *19*(2), p.431.
11. Zheng, Z., Ma, H., Lyu, M.R. and King, I., 2009, July. Wsrec: A collaborative filtering based web service recommender system. In *2009 IEEE International Conference on Web Services* (pp. 437-444). IEEE.
12. Geetha, G., Safa, M., Fancy, C. and Saranya, D., 2018, April. A hybrid approach using collaborative filtering and content based filtering for recommender system. In *Journal of Physics: Conference Series* (Vol. 1000, No. 1, p. 012101). IOP Publishing.
13. Gaw, F., 2022. Algorithmic logics and the construction of cultural taste of the Netflix Recommender System. *Media, Culture & Society*, *44*(4), pp.706-725.
14. Anwar, T. and Uma, V., 2019. A review of recommender system and related dimensions. *Data, engineering and applications*, pp.3-10.
15. Anwar, T. and Uma, V., 2019. A review of recommender system and related dimensions. *Data, engineering and applications*, pp.3-10.
16. Geetha, G., Safa, M., Fancy, C. and Saranya, D., 2018, April. A hybrid approach using collaborative filtering and content based filtering for recommender system. In *Journal of Physics: Conference Series* (Vol. 1000, No. 1, p. 012101). IOP Publishing.
17. Afoudi, Y., Lazaar, M. and Al Achhab, M., 2021. Hybrid recommendation system combined content-based filtering and collaborative prediction using artificial neural network. *Simulation Modelling Practice and Theory*, *113*, p.102375.
18. Lika, B., Kolomvatsos, K. and Hadjiefthymiades, S., 2014. Facing the cold start problem in recommender systems. *Expert systems with applications*, *41*(4), pp.2065-2073.

19. Natarajan, S., Vairavasundaram, S., Natarajan, S. and Gandomi, A.H., 2020. Resolving data sparsity and cold start problem in collaborative filtering recommender system using linked open data. *Expert Systems with Applications*, *149*, p.113248.
20. de Graaff, V., van de Venis, A., van Keulen, M. and Rolf, A., 2015, September. Generic knowledge-based Analysis of Social Media for Recommendations. In *CBRecSys@ RecSys* (pp. 22-29).
21. Chen, L.C., Kuo, P.J. and Liao, I.E., 2015. Ontology-based library recommender system using MapReduce. *Cluster Computing*, *18*(1), pp.113-121.
22. Ma, C., Gong, W., Hernández-Lobato, J.M., Koenigstein, N., Nowozin, S. and Zhang, C., 2018. Partial VAE for hybrid recommender system. In *NIPS Workshop on Bayesian Deep Learning* (Vol. 2018).
23. Gräßer, F., Beckert, S., Küster, D., Schmitt, J., Abraham, S., Malberg, H. and Zaunseder, S., 2017. Therapy decision support based on recommender system methods. *Journal of healthcare engineering*, *2017*.
24. Hu, Y., Chapman, A., Wen, G. and Hall, D.W., 2022. What can knowledge bring to machine learning?—a survey of low-shot learning for structured data. *ACM Transactions on Intelligent Systems and Technology (TIST)*, *13*(3), pp.1-45.
25. Dayan, P. and Balleine, B.W., 2002. Reward, motivation, and reinforcement learning. *Neuron*, *36*(2), pp.285-298.
26. Ludvig, E.A., Bellemare, M.G. and Pearson, K.G., 2011. A primer on reinforcement learning in the brain: Psychological, computational, and neural perspectives. In *Computational neuroscience for advancing artificial intelligence: Models, methods and applications* (pp. 111-144). IGI Global.
27. Even-Dar, E., Mannor, S., Mansour, Y. and Mahadevan, S., 2006. Action Elimination and Stopping Conditions for the Multi-Armed Bandit and Reinforcement Learning Problems. *Journal of machine learning research*, *7*(6).
28. Koulouriotis, D.E. and Xanthopoulos, A., 2008. Reinforcement learning and evolutionary algorithms for non-stationary multi-armed bandit problems. *Applied Mathematics and Computation*, *196*(2), pp.913-922.
29. Wang, K., Liu, Q. and Chen, L., 2012. Optimality of greedy policy for a class of standard reward function of restless multi-armed bandit problem. *IET signal processing*, *6*(6), pp.584-593.
30. Krishnamurthy, V., Wahlberg, B. and Lingelbach, F., 2005. A value iteration algorithm for partially observed markov decision process multi-armed bandits. *Math. of Oper. Res*, pp.133-152.
31. Rosman, B., Hawasly, M. and Ramamoorthy, S., 2016. Bayesian policy reuse. *Machine Learning*, *104*(1), pp.99-127.
32. Agarwal, S., Rodriguez, M.A. and Buyya, R., 2021, May. A reinforcement learning approach to reduce serverless function cold start frequency. In *2021 IEEE/ACM 21st International Symposium on Cluster, Cloud and Internet Computing (CCGrid)* (pp. 797-803). IEEE.
33. Tabatabaei, S.A., Hoogendoorn, M. and Halteren, A.V., 2018, October. Narrowing reinforcement learning: Overcoming the cold start problem for

16
personalized health interventions. In *International Conference on Principles and Practice of Multi-Agent Systems* (pp. 312-327). Springer, Cham.

34. Zou, L., Xia, L., Du, P., Zhang, Z., Bai, T., Liu, W., Nie, J.Y. and Yin, D., 2020, January. Pseudo Dyna-Q: A reinforcement learning framework for interactive recommendation. In *Proceedings of the 13th International Conference on Web Search and Data Mining* (pp. 816-824).
35. MacGregor, K. Access, retention and student success–A global view. *Student Affairs and Services in Higher Education: Global Foundations, Issues, and Best Practices Third Edition*, 107.
36. Rajagopalan, R. and Midgley, G., 2015. Knowing differently in systemic intervention. *Systems Research and Behavioral Science*, *32*(5), pp.546-561.
37. Burns, M.K., Deno, S.L. and Jimerson, S.R., 2007. Toward a unified response-to-intervention model. In *Handbook of response to intervention* (pp. 428-440). Springer, Boston, MA.https://link.springer.com/content/pdf/10.1007%2F978-0-387-49053-3.pdf
38. Zhao, C., Watanabe, K., Yang, B., & Hirate, Y. (2018, June). Fast converging multi-armed bandit optimization using probabilistic graphical model. In *Pacific-Asia Conference on Knowledge Discovery and Data Mining* (pp. 115-127). Springer, Cham. [9]  Leitner, P., Khalil, M. and Ebner, M., 2017. Learning analytics in higher education—a literature review. *Learning analytics: Fundaments, applications, and trends*, pp.1-23.
39. Gupta, S. (2020). Higher Education Management, Policies and Strategies. *Journal of Business Management & Quality Assurance (e ISSN 2456-9291)*, *1*(1), 5-11.
40. Kuh, G. D., & Kinzie, J. (2018). What really makes a "high-impact" practice high impact. *Inside Higher Ed*.
41. Organ, D., Dick, S., Hurley, C., Heavin, C., Linehan, C., Dockray, S., ... & Byrne, M. (2018). A systematic review of user-centred design practices in illicit substance use interventions for higher education students. In *European Conference on Information Systems 2018: Beyond Digitization-Facets of Socio-Technical Change*. AIS Electronic Library (AISeL)..
42. Cupák, A., Fessler, P., Silgoner, M., & Ulbrich, E. (2021). Exploring differences in financial literacy across countries: the role of individual characteristics and institutions. *Social Indicators Research*, 1-30.
43. Lacave, C., Molina, A. I., & Cruz-Lemus, J. A. (2018). Learning Analytics to identify dropout factors of Computer Science studies through Bayesian networks. *Behaviour & Information Technology*, *37*(10-11), 993-1007. *Fundaments, applications, and trends*, pp.1-23.
44. Scanagatta, M., Salmerón, A., & Stella, F. (2019). A survey on Bayesian network structure learning from data. *Progress in Artificial Intelligence*, *8*(4), 425-439.
45. Zhao, C., Watanabe, K., Yang, B., & Hirate, Y. (2018, June). Fast converging multi-armed bandit optimization using probabilistic graphical model. In *Pacific-Asia Conference on Knowledge Discovery and Data Mining* (pp. 115-127). Springer, Cham. [9]  Leitner, P., Khalil, M. and Ebner, M., 2017.

Learning analytics in higher education—a literature review. *Learning analytics: Fundaments, applications, and trends*, pp.1-23